\documentclass[lettersize,journal]{IEEEtran}
\usepackage{amsmath,amsfonts}
\usepackage{algorithmic}
\usepackage{algorithm}
\usepackage{array}
\usepackage[caption=false,font=normalsize,labelfont=sf,textfont=sf]{subfig}
\usepackage{textcomp}
\usepackage{stfloats}
\usepackage{url}
\usepackage{verbatim}
\usepackage{graphicx}
\usepackage{cite}
\usepackage{color, xcolor}
\usepackage{multirow}
\usepackage{hyperref}
\hypersetup{hidelinks}
\begin{document}
%

\title{TFKAN: Time-Frequency KAN for Long-Term Time Series Forecasting}

\author{Xiaoyan Kui, Canwei Liu, Qinsong Li$^*$, Zhipeng Hu, Yangyang Shi, Weixin Si, and Beiji Zou
\thanks{This work was supported in part by the National Natural Science Foundation of China (Nos. U22A2034, 62177047, 62302530), High Caliber Foreign Experts Introduction Plan funded by MOST, Key Research and Development Programs of Department of Science and Technology of Hunan Province (No. 2024JK2135), Major Program from Xiangjiang Laboratory (No. 23XJ02005), the Scientific Research Fund of Hunan Provincial Education Department (No. 24A0018), Hunan Provincial Natural Science Foundation (No. 2023JJ40769) and Central South University Research Programme of Advanced Interdisciplinary Studies (No. 2023QYJC020). (Corresponding author: Qinsong Li)}
\thanks{Xiaoyan Kui, Canwei Liu, Zhipeng Hu, Yangyang Shi, and Beiji Zou are with the School of Computer Science and Engineering, Central South University, Changsha, 410083, China (e-mail: xykui@csu.edu.cn, canwei\_liu@163.com, 244701046@csu.edu.cn, yyshi806@csu.edu.cn, bjzou@csu.edu.cn).}
\thanks{Qinsong Li is with the Big Data Institute, Central South University, Changsha, 410083, China (e-mail: qinsli.cg@csu.edu.cn).}
\thanks{Weixin Si is with the Shenzhen Institute of Advanced Technology, Chinese Academy of Sciences, Shenzhen, 518055, China (e-mail: wx.si@siat.ac.cn).}}

\maketitle

\begin{abstract}
Kolmogorov-Arnold Networks (KANs) are highly effective in long-term time series forecasting due to their ability to efficiently represent nonlinear relationships and exhibit local plasticity. However, prior research on KANs has predominantly focused on the time domain, neglecting the potential of the frequency domain. The frequency domain of time series data reveals recurring patterns and periodic behaviors, which complement the temporal information captured in the time domain. To address this gap, we explore the application of KANs in the frequency domain for long-term time series forecasting. By leveraging KANs’ adaptive activation functions and their comprehensive representation of signals in the frequency domain, we can more effectively learn global dependencies and periodic patterns. To integrate information from both time and frequency domains, we propose the \textbf{T}ime-\textbf{F}requency KAN (TFKAN). TFKAN employs a dual-branch architecture that independently processes features from each domain, ensuring that the distinct characteristics of each domain are fully utilized without interference. Additionally, to account for the heterogeneity between domains, we introduce a dimension-adjustment strategy that selectively upscales only in the frequency domain, enhancing efficiency while capturing richer frequency information. Experimental results demonstrate that TFKAN consistently outperforms state-of-the-art (SOTA) methods across multiple datasets. The code is available at \url{https://github.com/LcWave/TFKAN}.
\end{abstract}


\section{Introduction}
\IEEEPARstart{T}{ime} series forecasting (TSF) is crucial in various domains, such as financial modeling, healthcare diagnostics, and weather forecasting \cite{i1,jia2024witran, lim2021time, kim2025comprehensive}. Accurate long-term time series forecasting (LTSF) provides greater convenience, enabling more informed planning and decision-making \cite{su2025systematic, chen2023long}. Unlike short-term forecasting, long-term forecasting cannot rely solely on recent temporal information, such as trends, in the time domain. In other words, stock prices do not follow patterns from just the last few days, they must capture stable periodicity within the time series. Therefore, most models leverage the Transformer's ability \cite{wen2023transformers, zhou2021informer, wu2021autoformer, Yuqietal-2023-PatchTST} to model long-term dependencies for LTSF tasks.

Kolmogorov-Arnold Networks (KANs) \cite{liu2024kan} have recently emerged as a promising approach for LTSF due to their efficient nonlinear representation capabilities and local plasticity \cite{xu2024kolmogorov, vaca2024kolmogorov}. The local plasticity of B-Spline enables KAN to model complex patterns while preserving previously learned knowledge, making them particularly suitable for the LTSF. Unlike traditional neural network models, such as Multilayer Perceptrons (MLPs) \cite{yi2024frequency} and Transformer \cite{wu2021autoformer}, KAN mitigates the problem of catastrophic forgetting by leveraging local spline-based parametrizations. This unique feature allows KAN to adapt to new information without disrupting existing representations, enhancing its robustness for long-term sequential learning scenarios.

The frequency domain complements the time domain by providing insights into recurring cycles, periodicities, and spectral distributions that are critical for understanding long-term patterns \cite{cao2020spectral}. Recent studies have shown that periodic patterns are often more salient and interpretable in the frequency domain \cite{yi2024frequency,Zeng2022AreTE}.
Despite these advantages, most existing works on KAN have been restricted to time-domain modeling \cite{han2024kan4tsf, genet2024tkan, genet2024temporal, bhattacharya2024zero, lee2024hippo}. Although the recent TimeKAN \cite{huang2025timekan} employs FFT/IFFT to extract frequency components, the KAN modules themselves are applied only in the time domain. This limits their capacity to directly capture frequency-localized patterns. To the best of our knowledge, no prior work has explicitly applied KAN in the frequency domain. This leaves an open gap in leveraging KAN’s potential to jointly capture temporal dependencies and frequency-specific patterns, especially for LTSF tasks.

\begin{table}[h!t]
    \centering
    \caption{Synthetic functions used in toy experiments. Each function consists of a different combination of sine and cosine terms, representing varying frequency patterns.}
    \begin{tabular}{cc}
    \hline
    \textbf{ID} & \textbf{Target Function} \\ \hline
        F1 & $\sin(2\pi x) + 0.5\cos(4\pi x)$ \\
        F2 & $\sin(4\pi x) + 0.3\cos(8\pi x)$ \\
        F3 & $\sin(2\pi x) + \cos(6\pi x) + 0.3\cos(10\pi x)$ \\
        F4 & $\sin(2\pi x) + \sin(4\pi x + \frac{\pi}{3}) + \cos(6\pi x)$ \\ 
        \hline
    \end{tabular}
    \label{tab: toy_functions}
    \vspace{-1em}
\end{table}

\begin{figure}[h!t]
    \centering
    \includegraphics[width=1\linewidth]{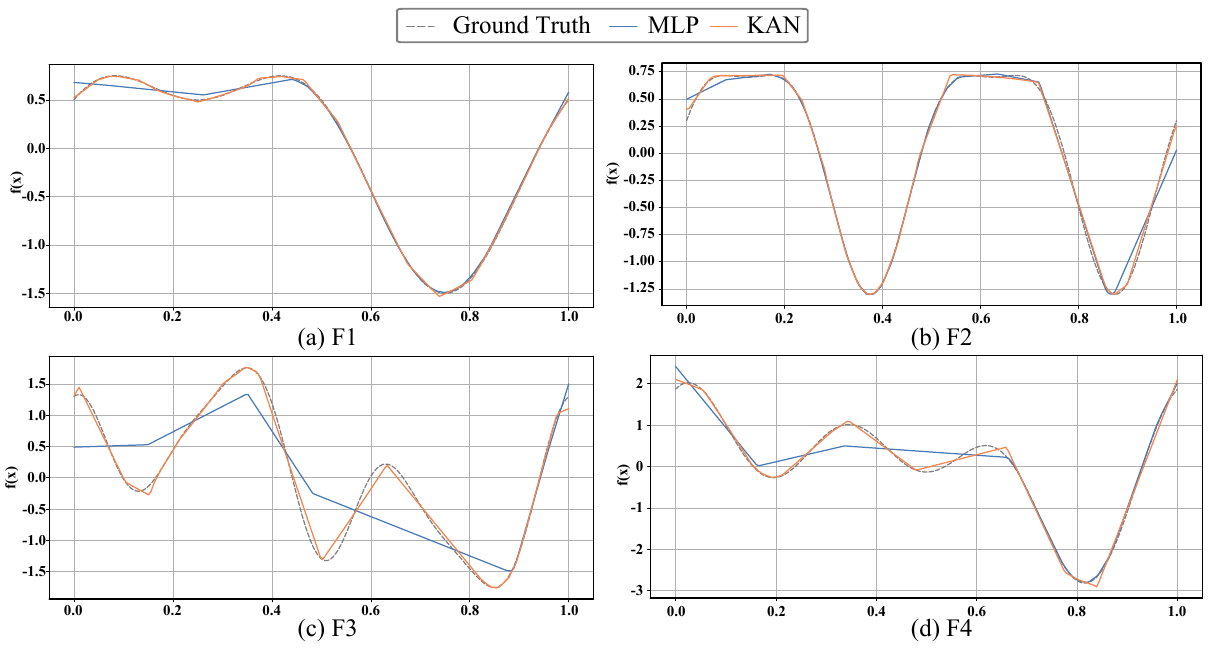}
    \caption{Comparison of MLP and KAN in approximating periodic functions (F1–F4). KAN consistently achieves smoother and more accurate reconstructions, especially under high-frequency and phase-shifted conditions.}
    \label{fig: toy_experiments}
    \vspace{-1em}
\end{figure}

To gauge whether the spline‐based activations of KANs are advantageous for modelling periodic structure, we first perform a controlled function approximation study in the time domain. In detail, we synthesise four target functions (Table \ref{tab: toy_functions}) by summing sinusoids of different frequencies and phases. These signals contain multiple harmonic components and therefore mimic the mixed periodicities often encountered in real LTSF tasks. Then, we compare a two‑layer ReLU MLP against a KAN with grid size $s=2$ and spline order $k=1$, training both to regress each $f_i(x)$ on $x \in [0,1]$. Results in Fig. \ref{fig: toy_experiments} show that KAN consistently yields smoother and more accurate reconstructions. Because a Fourier transform maps multi‑harmonic signals to sparse, localised spectral peaks, the observed advantage suggests that KAN’s B‑spline bases naturally adapt around those peaks. This motivates us to place KAN modules directly in the frequency domain, where periodic information is explicit and global dependencies can be captured with greater interpretability.

Based on these, we propose the frequency-domain KAN. To the best of our knowledge, this is the first attempt to directly apply KAN in the frequency domain for time series forecasting. By leveraging KANs’ adaptive activation functions and their comprehensive representation of signals in the frequency domain, we can more effectively learn global dependencies and periodic patterns. Furthermore, to integrate information from both time and frequency domains, we introduce the \textbf{T}ime-\textbf{F}requency \textbf{K}olmogorov-\textbf{A}rnold \textbf{N}etworks, named TFKAN. This architecture features a dual-branch structure that independently processes features from each domain, ensuring that the distinct characteristics of each domain are fully utilized without interference. By designing specialized KANs in each branch, TFKAN optimizes feature extraction for both domains, enabling effective capture of domain-specific features. Additionally, a dimension-adjustment strategy is implemented to address the heterogeneity between the time and frequency domains. Specifically, downscaling in the time domain adapts temporal features for efficient processing, while upscaling in the frequency domain highlights periodic patterns for better representation, which ensures efficient utilization of information from both domains.

The contributions of this paper are summarized as follows:
\begin{itemize}
    \item We propose the frequency-domain KAN, a novel approach that enables the model to capture prominent periodic patterns in the frequency domain. To the best of our knowledge, this is the first work to directly apply KAN in the frequency domain for time series forecasting.
    \item We introduce a dual-branch architecture TFKAN that independently processes features from the time and frequency domains. This design ensures the full utilization of the unique characteristics of each domain while preventing interference between them.
    \item We propose a dimension-adjustment strategy to address the heterogeneity between the time and frequency domains. This strategy selectively upscales only in the frequency domain, enhancing computational efficiency while capturing richer frequency information.
    \item Through extensive experiments on seven time-series datasets, we demonstrate that TFKAN outperforms eight SOTA methods, underscoring its superior forecasting capabilities.
\end{itemize}

\section{Related Work}\label{sec:related_work}

Recent advancements in long-term time series forecasting (LTSF) can be broadly categorized into three modeling paradigms based on their primary representation domain: \textbf{time-based}, \textbf{frequency-based}, and \textbf{hybrid} approaches. This section briefly reviews each category.

\textbf{Time-Based Models.}  
Time-domain models directly model temporal dynamics using linear projections, MLPs, or attention mechanisms. Transformer-based methods such as LogTrans \cite{li2019enhancing}, TFT \cite{lim2021temporal}, and Informer \cite{zhou2021informer} adapt attention-based architectures to model long-range dependencies. More recent developments, like PatchTST \cite{Yuqietal-2023-PatchTST}, PETformer \cite{lin2024petformer}, and Crossformer \cite{zhang2023crossformer}, improve performance and efficiency by partitioning inputs into patches.
Meanwhile, lightweight linear and MLP-based models offer faster inference with fewer parameters. LTSF-Linear \cite{Zeng2022AreTE} demonstrates the surprising effectiveness of a single-layer linear model, inspiring subsequent works like LightTS \cite{zhang2022less}, TiDE \cite{das2023long}, MTS-Mixers \cite{li2023mts}, TimeMixer \cite{wang2024timemixer}, and HDMixer \cite{huang2024hdmixer}. WormKAN \cite{xu2024kan} and RMoK \cite{han2024kan4tsf} are KAN-based methods designed for concept drift and variable-specific modeling, respectively. TKAN \cite{genet2024tkan} as a recurrent KAN architecture marrying KAN with LSTM-like memory. And TKAT \cite{genet2024temporal} is an encoder–decoder architecture that injects TKAN layers into a Transformer framework. C-KAN \cite{livieris2024c} employs convolutional layers to extract local temporal patterns before feeding them into a KAN layer. However, neither leverages frequency features.

\textbf{Frequency-Based Models.}  
Frequency-domain models exploit frequency information to capture periodic patterns and global trends. FreTS \cite{yi2024frequency} enhances MLPs by applying frequency decomposition with energy compaction. FEDformer \cite{zhou2022fedformer} integrates Fourier-based convolution with decomposition mechanisms for better trend-seasonal modeling. FITS \cite{xu2024fits} operates entirely in the complex frequency domain, leveraging interpolation over complex value's components rather than processing raw time-domain sequences. Although these models benefit from frequency-domain representations, none of them perform function learning directly on the complex value's components using KAN.

\textbf{Hybrid Time-Frequency Models.}  
Hybrid approaches aim to combine the strengths of both domains. Autoformer \cite{wu2021autoformer} integrates auto-correlation and progressive decomposition for temporal and periodic pattern extraction.
TimeKAN \cite{huang2025timekan} incorporates KANs alongside FFT/IFFT operations but still applies KANs only in the time domain.
ATFNet \cite{ye2024atfnet} uses a dual-branch architecture to process both domains simultaneously, featuring an extended DFT and complex-valued attention. JTFT \cite{chen2024joint}, T-FIA \cite{yang2024t}, and TFMRN \cite{yan2024multi} further demonstrate the effectiveness of concurrent time-frequency modeling. These methods, however, rely on predefined transformations (e.g., DFT) or attention mechanisms, and do not incorporate learnable function approximation in the frequency domain.

\begin{figure*}[htbp]
    \centering
    \includegraphics[width=1\linewidth]{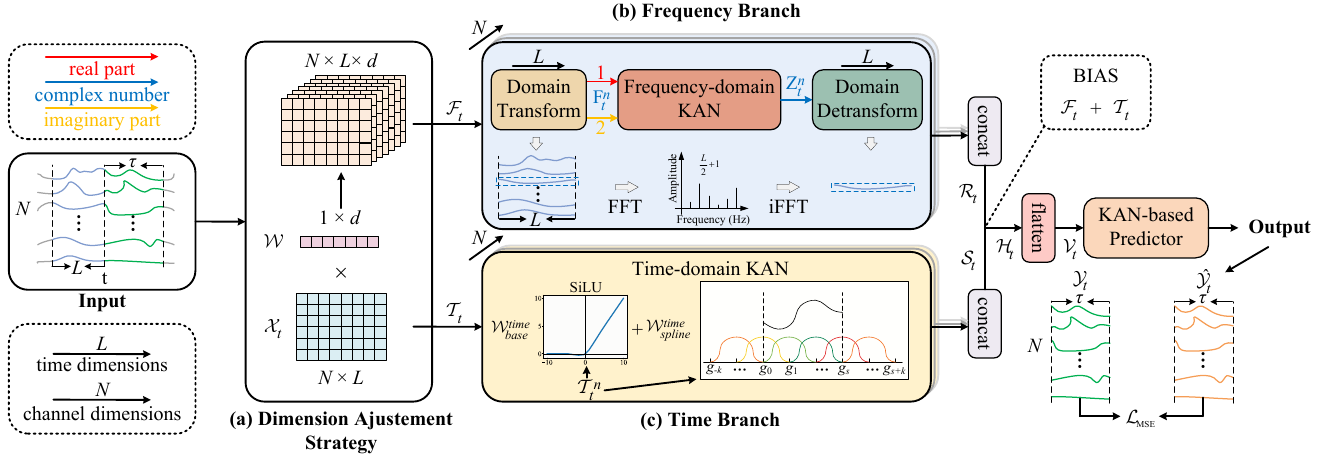}
    \caption{The overall architecture of TFKAN: (a) Dimension Adjustment Strategy is used to prepare the data $\mathcal{F}_t$ and $\mathcal{T}_t$ for the Frequency and Time Branches, separately. (b) The Frequency Branch extracts periodic patterns from the frequency domain using the Frequency-domain KAN. (c) The Time Branch captures temporal dependencies from the time domain with the Time-Domain KAN.}
    \label{fig:overall architecture}
\end{figure*}

\section{Methods}\label{sec:methods}

In this section, the details of the proposed TFKAN framework are presented. Firstly, the LTSF problem is formally defined in Section \ref{sec: problem}. Secondly, the prerequisites of Kolmogorov-Arnold Networks (KAN) are provided in Section \ref{sec: prerequisites}. Finally, Section \ref{sec: dual-branch} introduces the dual-branch architecture of TFKAN, integrating Frequency-Domain KAN (FreqKAN), Time-Domain KAN (TimeKAN), and a KAN-based Predictor.

\subsection{Problem Definition} \label{sec: problem}

For LTSF, the historical data is represented as $ \mathcal{X} = [X_1, \ldots, X_T] \in \mathbb{R}^{N \times T} $, where $N$ denotes the number of variables (or features) and $T$ represents the number of time steps. Each $X_t \in \mathbb{R}^N$ contains the multivariate values of $N$ variables at time step $t$. A segment of the time series with a lookback window of length $L$ at timestamp $t$ is used as the model input denoted as $ \mathcal{X}_t = [X_{t-L+1}, X_{t-L+2}, \ldots, X_t] \in \mathbb{R}^{N \times L} $. The objective is to predict future values $ \mathcal{Y}_t = [X_{t+1}, \ldots, X_{t+\tau}] \in \mathbb{R}^{N \times \tau} $ over the next $\tau$ time steps, where $\tau$ represents the long-term forecast horizon. This is done using a forecasting model $f_\theta$, such that $\hat{\mathcal{Y}}_t = f_\theta(\mathcal{X}_t)$.

\subsection{Prerequisites} \label{sec: prerequisites}
KAN \cite{liu2024kan} is grounded in the Kolmogorov-Arnold Representation Theorem, which states that any multivariate continuous function can be represented as a finite combination of univariate continuous functions. KAN builds on this foundation by incorporating two key mechanisms: Base Transformation and B-Spline Transformation. These mechanisms work synergistically to capture complex relationships while maintaining local plasticity, enabling the model to adapt to new inputs without overwriting previously learned information.

\textbf{Base Transformation.} The base transformation captures primary patterns in the input data through a linear mapping followed by a nonlinear activation. It employs a learned weight matrix $\mathcal{W}_{base}$ and the SiLU activation function \cite{elfwing2018sigmoid}. This transformation is defined as:
\begin{equation}
        \mathbf{z}_{base} = \mathcal{W}_{base} \cdot {\rm{SiLU}}(x),
\end{equation}
where $x$ represents the input data, and $\mathbf{z}_{base}$ denotes the transformed output.

\textbf{B-Spline Transformation.} The B-spline transformation performs smooth interpolation between data points, allowing for flexible modeling of complex patterns. It leverages a uniformly spaced grid $\mathbf{G} \in \mathbb{R}^{s + 2 \times k + 1}$, where $s$ is the grid size, indicating the number of interpolation points, and $k$ is the spline order. This transformation is expressed as: 

\begin{equation}
    \mathbf{z}_{spline} = \sum_{i} c_i \cdot \mathbf{B}_i^k(x),
\end{equation}
where  $c_i$s are learnable weights. In this paper, these are denoted together with $\mathcal{W}_{spline}$. The $\mathbf{B}_i^k$ represents the $i$-th B-spline of degree $k$, and $\mathbf{z}_{spline}$ is the interpolated output. The grid $\mathbf{G} = [g_{-k}, \dots, g_0, g_1 \dots, g_{s + k}]$ is uniformly distributed over $[-1, 1]$, controlling the resolution of spline interpolation. The B-spline bases $\mathbf{B}_i^k(x)$ are recursively computed as:

\begin{equation}
\resizebox{.91\linewidth}{!}{$\begin{aligned}
    \mathbf{B}^0_i(x) &= \begin{cases} 
1 & \text{if } x \in [g_i, g_{i+1}) \\
0 & \text{otherwise}
\end{cases} \\
    \mathbf{B}^k_i(x) &= \frac{x - g_i}{g_{i+k} - g_i} \mathbf{B}^{k-1}_i(x) + \frac{g_{i+k+1} - x}{g_{i+k+1} - g_{i+1}} \mathbf{B}^{k-1}_{i+1}(x),
\end{aligned}$}
\end{equation}
where $\mathbf{B}_i^0(x)$ represents a 0-degree B-spline, and $\mathbf{B}_i^k(x)$ represents a $k$-degree B-spline. The index $i$ denotes the $i$-th spline base.

Finally, the combined KAN output is defined as:
\begin{equation}
    \mathbf{z} = \mathbf{z}_{base} + \mathbf{z}_{spline}.
\end{equation}

\subsection{TFKAN: Time-Frequency KAN }\label{sec: dual-branch}

Building upon the strengths of KAN described above, the dual-branch architecture of TFKAN is designed to handle the heterogeneity of time and frequency domains. The architecture leverages two distinct branches, a Frequency Branch and a Time Branch, to process frequency-domain and time-domain representations independently, ensuring optimized feature extraction in each domain. The architecture, illustrated in Fig. \ref{fig:overall architecture}, also incorporates a Dimension Adjustment Strategy and a KAN-based Predictor.

\textbf{Overview of Dual-Branch Workflow.}
At each time step $t$, the historical input $\mathcal{X}_t$ is first preprocessed by the Dimension Adjustment Strategy (Fig. \ref{fig:overall architecture} (a)) to prepare the data for the Frequency and Time Branches (Fig. \ref{fig:overall architecture} (b) and (c)). The Frequency Branch extracts spectral features using a DomainTransform operation and processes them using the Frequency-Domain KAN. After that, the output of the frequency-domain KAN is processed by a DomainDetransform operation. Meanwhile, the Time Branch directly captures temporal dependencies through the Time-Domain KAN. Finally, the outputs of both branches are integrated with a bias term and fed into the KAN-based Predictor to generate the final forecast.

\textbf{Dimension Adjustment Strategy.} To optimize feature extraction, the dimension adjustment strategy (Fig. \ref{fig:overall architecture} (a)) modifies the input data differently for the frequency and time branches. For the frequency branch, the historical data $\mathcal{X}_t \in \mathbb{R}^{N \times L}$ is multiplied by a learnable weight vector $\mathcal{W} \in \mathbb{R}^{1 \times d}$, producing hidden representations $\mathcal{F}_t \in \mathbb{R}^{N \times L \times d}$, which are enriched with frequency-specific information. In this paper, the embedding size $d$ is set as $128$. For the time branch, the original input remains unchanged as $\mathcal{T}_t \in \mathbb{R}^{N \times L}$ to preserve the temporal structure and efficient processing. The operations are formally defined as:

\begin{equation}
    \begin{aligned}
        {{\cal F}_t} &= {{\cal X}_t} \times {\cal W}\\
        {{\cal T}_t} &= {{\cal X}_t}.
    \end{aligned}
\end{equation}

\textbf{DomainTransform/Detransform.} The DomainTransform operation converts data from the time domain to the frequency domain using the Fast Fourier Transform (FFT). This operation decomposes the time signal into its frequency components. Conversely, the DomainDetransform operation employs the Inverse Fast Fourier Transform (IFFT) to map frequency-domain data back to the time domain. Specifically, the input of the Frequency Branch $\mathcal{F}_t$, which is treated as continuous data, is transformed into the frequency domain, $\mathbf{F}_t \in \mathbb{C}^{N \times (\frac{L}{2} +1) \times d}$, by:

\begin{small}
    \begin{equation}
    \resizebox{.91\linewidth}{!}{$\begin{aligned}
        \mathbf{F}_t(f) &= \int_{-\infty}^{\infty} \mathcal{F}_t(v) e^{-j 2 \pi f v} dv \\
        &= \int_{-\infty}^{\infty} \mathcal{F}_t(v) \cos(2 \pi f v) dv - j \int_{-\infty}^{\infty} \mathcal{F}_t(v) \sin(2 \pi f v) dv ,
    \end{aligned}$} \label{eq:domain transform}
    \end{equation}
\end{small}where $f$ denotes the frequency variable, $v$ denotes the integral variable, and $j = \sqrt{-1}$ is the imaginary unit. Additionally, the integral $\int_{-\infty}^{\infty} \mathcal{F}_t(v) \cos(2 \pi f v) dv$ represents the real part of $\mathbf{F}_t$, denoted as $Re(\mathbf{F}_t)$. Similarly, the integral $\int_{-\infty}^{\infty} \mathcal{F}_t(v) \sin(2 \pi f v) dv$ represents the imaginary part of $\mathbf{F}_t$, denoted as $Im(\mathbf{F}_t)$. In other words, the frequency-domain data consists of $\cos$ and $\sin$ waves with varying frequencies and phases, and their magnitudes represent the corresponding amplitudes. 

After completing all operations in the frequency domain, the frequency data $\mathbf{F}_t$ is transformed back into the time domain as follows:

\begin{small}
    \begin{equation}
    \resizebox{.91\linewidth}{!}{$\begin{aligned}
        \mathcal{R}_t(v) &= \int_{-\infty}^{\infty} \mathbf{F}_t(f) e^{j 2 \pi f v} df \\
        &= \int_{-\infty}^{\infty} \left[ Re(\mathbf{F}_t(f)) + j Im(\mathbf{F}_t(f)) \right] e^{j 2 \pi f v} df,
    \end{aligned}$} \label{eq:domain detransform}
    \end{equation}
\end{small}where $f$ denotes the integral variable. Notably, the DomainTransform and DomainDetransform operations are applied only to the frequency branch.

\textbf{Frequency Branch.}
The Frequency Branch (Fig. \ref{fig:overall architecture} (b)) is designed to learn frequency features and periodic patterns in the frequency domain.  It operates channel-wise, with each input channel processed independently. Taking the $n$-th channel of $\mathcal{F}_t$, denoted as $\mathcal{F}_t^n \in \mathbb{R}^{L \times d}$, as an example. The processing steps of the Frequency Branch are as follows:

\begin{equation}
    \begin{aligned}
        \mathbf{F}_t^n &= \mathrm{DomainTransform}_{(temp)} (\mathcal{F}_t^n) \\
        \mathbf{Z}_t^n &= \mathrm{FreqKAN} (\mathbf{F}_t^n, \mathcal{W}^{freq}_{base}, \mathcal{W}^{freq}_{spline}) \\
        \mathcal{R}_t^n &= \mathrm{DomainDetransform}_{(temp)} (\mathbf{Z}_t^n),
    \end{aligned} \label{eq:frequency branch}
\end{equation}
where $\mathbf{F}_t^n \in \mathbb{C}^{1 \times (\frac{L}{2}+1) \times d}$ represents the frequency-domain data obtained through the DomainTransform. $\mathbf{Z}_t^n \in \mathbb{C}^{1 \times (\frac{L}{2}+1) \times d}$ refers to the processed data from the Frequency-Domain KAN. Additionally, the matrices $\mathcal{W}^{freq}_{base} \in \mathbb{R}^{d \times d}$ and $\mathcal{W}^{freq}_{spline} \in \mathbb{R}^{d \times d \times (s+k)}$ represent the weight matrices for the base function $\text{SiLU}$ and the B-spline function, respectively, where the $s$ denotes grid size and the $k$ denotes the spline order. Finally, $\mathcal{R}_t^n$ is the reconstructed time-domain data obtained after applying the DomainDetransform. The operations $\mathrm{DomainTransform}_{(temp)}$ and $\mathrm{DomainDetransform}_{(temp)}$ are performed along the temporal dimension, i.e., the $L$-dimension. 

The Frequency-Domain KAN ($\text{FreqKAN}(\cdot)$) is a two-layer KAN, the network processes the real and imaginary components of $\mathbf{F}_t^n \in \mathbb{C}^{1 \times (\frac{L}{2} + 1) \times d}$ separately. More specifically, the real part $Re(\mathbf{F}_t^n) \in \mathbb{R}^{1 \times (\frac{L}{2} + 1) \times d}$ is firstly input into the two-layer KAN to get the real output $Re(\mathbf{Z}^n_t)$. Then, the imaginary part $Im(\mathbf{F}_t^n) \in \mathbb{C}^{1 \times (\frac{L}{2} + 1) \times d}$ is input into the network to get the imaginary output $Im(\mathbf{Z}^n_t)$. Due to the local plasticity of KAN, the features captured from the real and imaginary parts are effectively integrated.  Using the same KAN network for both the real and imaginary parts ensures consistent feature learning and parameter sharing for meaningful signal reconstruction. Finally, the real and imaginary outputs are combined to form the final complex-valued representation:

\begin{equation}
    \mathbf{Z}^n_t = Re(\mathbf{Z}^n_t) + jIm(\mathbf{Z}^n_t),
\end{equation}
where $\mathbf{Z}^n_t \in \mathbb{C}^{1 \times (\frac{L}{2}+1) \times d}$.

\textbf{Time Branch.}
The Time Branch (Fig. \ref{fig:overall architecture} (c)) is designed to capture temporal dependencies and trends directly from the time-domain data. Unlike the Frequency Branch, no transformation is applied to the input data before it is processed by the Time-Domain KAN, ensuring efficiency. The input data $\mathcal{T}_t \in \mathbb{R}^{N \times L}$ is passed directly to the Time-Domain KAN, which produces temporal representations $\mathcal{S}_t \in \mathbb{R}^{N \times L}$. The operations within the Time Branch also operate on the channel dimension. Taking the $n$-th channel of $\mathcal{T}_t$, denoted as $\mathcal{T}_t^n \in \mathbb{R}^{1 \times L}$, as an example, the process can be formally defined as:

\begin{equation}
    \mathcal{S}_t^n = \mathrm{TimeKAN}(\mathcal{T}_t^n, \mathcal{W}^{time}_{base}, \mathcal{W}^{time}_{spline}),
\end{equation}
where $\mathcal{S}_t^n \in \mathbb{R}^{1 \times L}$ refers to the processed data from the Time-Domain KAN. The weight matrices $\mathcal{W}^{time}_{base} \in \mathbb{R}^{L \times L}$ and $\mathcal{W}^{time}_{spline} \in \mathbb{R}^{L \times L \times (s+k)}$ are the learnable parameters in the Time-Domain KAN, respectively.

 The Time-Domain KAN ($\text{TimeKAN}(\cdot)$) processes inputs to capture temporal features and dependencies inherent in time series data. Its structure is similar to that of the Frequency-Domain KAN but processes inputs at one time. 

\textbf{Long-term Time Series Forecasting.} To combine the information from both the Frequency and Time branches, the outputs of these branches are integrated with a bias term. The process is mathematically formulated as follows:

\begin{equation}
\begin{aligned}
    \text{BIAS} &= \mathbf{F}^n_t + \mathcal{T}^n_t \\
    \mathcal{H}_t^n &= \mathcal{R}^n_t + \mathcal{S}^n_t + \text{BIAS},
\end{aligned}
\end{equation}
where $\mathcal{H}_t^n \in \mathbb{R}^{1 \times L \times d}$ denotes the final integrated hidden representation used for forecasting. Additionally, a broadcast mechanism is employed to compute $\text{BIAS} \in \mathbb{R}^{1 \times L \times d}$. Subsequently, $\mathcal{H}_t^n$ is fed into a KAN-based Predictor, to generate the prediction $\hat{\mathcal{Y}}_t^n \in \mathbb{R}^{1 \times \tau}$. 

The structure of the KAN-based Predictor is also similar to the Frequency-Domain KAN. Firstly, the input hidden representation $\mathcal{H}^n_t \in \mathbb{R}^{1 \times L \times d}$ is reshaped into a flattened vector $\mathcal{V}^n_t \in \mathbb{R}^{1 \times (L \cdot d)}$. Subsequently, the reshaped input is fed into a KAN with an input size of $L \cdot d$ and an output size of $\tau$. These operations are formulated as:

\begin{equation}
    \hat{\mathcal{Y}}_t^n = \text{KAN}(\mathcal{V}^n_t, \mathcal{W}_{base}^{pre}, \mathcal{W}_{spline}^{pre}).
\end{equation}

Finally, the prediction $\hat{\mathcal{Y}}_t^n$ is reconstructed against the ground truth $\mathcal{Y}^n_t \in \mathbb{R}^{1 \times \tau}$ using the Mean Squared Error (MSE) loss, which is computed as follows:

\begin{equation}
     \mathcal{L}_\text{MSE} = \frac{1}{\tau} \sum_{i=1}^{\tau} \left( \hat{\mathcal{Y}}_t^n - \mathcal{Y}_t^n[i] \right)^2, 
\end{equation}
where $i$ represents the $i$-th variable along the $\tau$-dimension.

\begin{table*}[htbp]
    \centering
    \caption{
    Summary of baseline models used for comparison. Methods are categorized into time-domain, frequency-domain, and hybrid architectures. Hybrid models combine both time and frequency domain representations. All source codes are publicly available.
    }
    \renewcommand{\arraystretch}{0.9}
    \begin{tabular}{l l p{21em} l}
    \hline
    \textbf{Method} & \textbf{Type} & \textbf{Full Name} & \textbf{Source Code} \\
    \hline
    TimeKAN \cite{huang2025timekan} & Hybrid & KAN-Based Frequency Decomposition Architecture for Long-Term Time Series Forecasting & \href{https://github.com/huangst21/TimeKAN}{Code} \\
    ATFNet \cite{ye2024atfnet} & Hybrid & Adaptive Time-Frequency Ensembled Network for Time Series Forecasting & \href{https://github.com/YHYHYHYHYHY/ATFNet}{Code} \\
    FreTS \cite{yi2024frequency} & Frequency & Frequency-Domain MLPs for Enhanced Time Series Forecasting & \href{https://github.com/aikunyi/FreTS}{Code} \\
    LTSF-Linear \cite{Zeng2022AreTE} & Time & Simple Linear Baselines for Long-Term Time Series Forecasting & \href{https://github.com/cure-lab/LTSF-Linear}{Code} \\
    TSMixer \cite{chen2023tsmixer} & Time & All-MLP Architecture for Time Series Forecasting & \href{https://github.com/google-research/google-research/tree/master/tsmixer}{Code} \\
    FEDformer \cite{zhou2022fedformer} & Frequency & Frequency Enhanced Decomposed Transformer for Long-Term Forecasting & \href{https://github.com/MAZiqing/FEDformer}{Code} \\
    Informer \cite{zhou2021informer} & Time & Efficient Transformer with ProbSparse Attention for Long Series & \href{https://github.com/zhouhaoyi/Informer2020}{Code} \\
    Autoformer \cite{wu2021autoformer} & Hybrid & Decomposition Transformer with Auto-Correlation Mechanism & \href{https://github.com/thuml/Autoformer}{Code} \\
    \hline
    \end{tabular}
    \label{tab: baselines}
\end{table*}

\section{Experiments}\label{sec:experiments}
In this section, extensive experiments are presented on seven time series datasets to answer the following research questions (RQ):

\begin{itemize}
    \item \textbf{RQ1 (Accuracy):} Does the proposed TFKAN outperform SOTA methods across various scenarios?
    \item \textbf{RQ2 (Ablation):} Do key components of TFKAN contribute to the overall performance?
    \item \textbf{RQ3 (Sensitivity \& Effciency):} How does TFKAN perform under different hyperparameter configurations? How about its computational efficiency?
\end{itemize}

The RQ1, RQ2, and RQ3 will be answered in Sections \ref{sec: mainResults} and \ref{sec: ablation}, and \ref{sec: sensitivity}-\ref{sec: efficiency}, respectively.

\subsection{Experimental Settings}

\textbf{Datasets.} Seven time-series datasets from diverse domains, including electricity, medical, and meteorology, are selected for evaluation. The datasets are as follows: \textbf{1) ETT\footnote{\url{https://github.com/zhouhaoyi/ETDataset}}:} The selected ETT dataset from ETT consists of four subsets: ETTm1, ETTm2, ETTh1, and ETTh2. It contains data from two station electricity transformers, including load and oil temperature measurements. Each transformer dataset is recorded at two resolutions: 15 minutes (denoted as 'm') and 1 hour (denoted as 'h'). \textbf{2) Air\footnote{\url{https://archive.ics.uci.edu/dataset/360/air+quality}}:} This dataset records the hourly responses of a gas multisensor device deployed in an Italian city, along with gas concentration references measured by a certified analyzer. \textbf{3) Weather\footnote{\url{https://www.bgc-jena.mpg.de/wetter/}}:} Collected in 2020 from the Weather Station of the Max Planck Biogeochemistry Institute in Germany, this dataset includes 21 meteorological indicators such as humidity and air temperature. The data is sampled every 10 minutes. \textbf{4) ILI\footnote{\url{https://gis.cdc.gov/grasp/fluview/fluportaldashboard.html}}:} This dataset contains weekly records of influenza-like illness (ILI) patient data from the Centers for Disease Control and Prevention in the United States, spanning 2002 to 2021. It describes the ratio of patients with ILI to the total number of patients. To ensure consistency, all datasets are normalized to the range [0, 1] using min-max normalization. The datasets are split into training, validation, and test sets with a ratio of 7:2:1, except for the ILI dataset, which is divided into a 6:2:2 ratio due to its shorter sequence lengths.

\begin{table*}[h!t]
    \centering
    \caption{Comparison of TFKAN against eight SOTA baselines. The lookback window is fixed at $L=96$ for all datasets. For the ILI dataset, prediction lengths are $\tau \in \{24,36,48,60\}$, while for the remaining datasets, $\tau \in \{96,192,336,720\}$. The best results are \textbf{bolded}, and the second-best are \underline{underlined}. The bottom two rows report the number of times each method achieves the best or second-best performance across all experiments.}
    \resizebox{0.99\textwidth}{!}{
    \begin{tabular}{cc|cc|cc|cc|cc|cc|cc|cc|cc|cc}
    \hline
        \multicolumn{2}{c|}{Model} & \multicolumn{2}{c|}{TFKAN} & \multicolumn{2}{c|}{TimeKAN} & \multicolumn{2}{c|}{ATFNet} & \multicolumn{2}{c|}{FreTS} & \multicolumn{2}{c|}{LTSF-Linear} & \multicolumn{2}{c|}{TSMixer} & \multicolumn{2}{c|}{FEDformer} & \multicolumn{2}{c|}{Informer} & \multicolumn{2}{c}{Autoformer} \\
        \multicolumn{2}{c|}{Metrics} & MAE & RMSE & MAE & RMSE & MAE & RMSE & MAE & RMSE & MAE & RMSE & MAE & RMSE & MAE & RMSE & MAE & RMSE & MAE & RMSE \\ \hline
        \multirow{4}*{\rotatebox{90}{ILI}} & 24 & \textbf{0.146}  & \textbf{0.211} & 0.151 & 0.223 & \underline{0.149} & \underline{0.216} & 0.157 & 0.228 & 0.167 & 0.236 & 0.231 & 0.306 & 0.184 & 0.251 & 0.241 & 0.344 & 0.205 & 0.273 \\ 
        ~ & 36 & \textbf{0.137}  & \textbf{0.206} & 0.146 & 0.216 & \underline{0.138} & \underline{0.209} & 0.165 & 0.229 & 0.160 & 0.226 & 0.238 & 0.318 & 0.180 & 0.250 & 0.274 & 0.383 & 0.204 & 0.274 \\ 
        ~ & 48 & \textbf{0.139}  & \textbf{0.206} & \underline{0.147} & \underline{0.216} & 0.147 & \underline{0.217} & 0.166 & 0.230 & 0.161 & 0.226 & 0.251 & 0.340 & 0.186 & 0.257 & 0.274 & 0.384 & 0.200 & 0.270 \\ 
        ~ & 60 & \textbf{0.149}  & \textbf{0.215} & 0.159 & 0.227 & \underline{0.150} & \underline{0.220} & 0.166 & 0.228 & 0.158 & 0.222 & 0.273 & 0.367 & 0.193 & 0.265 & 0.289 & 0.404 & 0.208 & 0.280 \\  \hline
        \multirow{4}*{\rotatebox{90}{ETTh1}} & 96 & \textbf{0.061}  & \textbf{0.088} & \underline{0.063} & 0.092 & 0.064 & 0.092 & \underline{0.063} & 0.091 & \underline{0.063} & 0.091 & 0.077 & 0.107 & \underline{0.063} & \underline{0.089} & 0.082 & 0.108 & 0.074 & 0.103 \\
        ~ & 192 & \textbf{0.067} & \textbf{0.093} & \textbf{0.067} & 0.097 & 0.073 & 0.101 & \underline{0.069} & \underline{0.096} & \textbf{0.067} & \underline{0.096} & 0.089 & 0.121 & \textbf{0.067} & \underline{0.096} & 0.115 & 0.146 & 0.078 & 0.109 \\ 
        ~ & 336 & \underline{0.071} & \textbf{0.098} & \textbf{0.069} & \underline{0.099} & 0.086 & 0.112 & 0.073 & 0.100 & \underline{0.071} & \underline{0.099} & 0.101 & 0.134 & 0.078 & 0.110 & 0.124 & 0.156 & 0.081 & 0.112 \\ 
        ~ & 720 & \underline{0.082} & \textbf{0.109} & \textbf{0.080} & \underline{0.110} & 0.116 & 0.150 & 0.084 & 0.111 & 0.083 & 0.111 & 0.119 & 0.150 & 0.086 & 0.117 & 0.125 & 0.159 & 0.086 & 0.119 \\ \hline
        \multirow{4}*{\rotatebox{90}{ETTh2}} & 96 & \underline{0.036} & \underline{0.052} & 0.045 & 0.068 & \textbf{0.035} & \textbf{0.051} & 0.038 & \underline{0.052} & \underline{0.036} & \textbf{0.051} & 0.090 & 0.114 & 0.052 & 0.076 & 0.051 & 0.067 & 0.052 & 0.075 \\ 
        ~ & 192 & \textbf{0.040} & \textbf{0.057} & 0.052 & 0.077 & 0.046 & 0.060 & 0.043 & \underline{0.059} & \underline{0.041} & \textbf{0.057} & 0.105 & 0.132 & 0.059 & 0.084 & 0.048 & 0.064 & 0.061 & 0.085 \\ 
        ~ & 336 & \textbf{0.041}  & \textbf{0.058} & 0.058 & 0.083 & 0.067 & 0.087 & 0.044 & \underline{0.060} & \textbf{0.042} & \textbf{0.058} & 0.117 & 0.148 & 0.068 & 0.093 & 0.060 & 0.079 & 0.065 & 0.089 \\ 
        ~ & 720 & \textbf{0.047} & \textbf{0.065} & 0.062 & 0.088 & 0.072 & 0.099 & 0.067 & 0.089 & \underline{0.053} & \underline{0.070} & 0.105 & 0.129 & 0.069 & 0.094 & 0.085 & 0.106 & 0.066 & 0.090 \\ \hline
        \multirow{4}*{\rotatebox{90}{ETTm1}} & 96 & \textbf{0.054} & \textbf{0.079} & \underline{0.055} & 0.082 & 0.056 & 0.082 & \textbf{0.054} & \textbf{0.079} & \underline{0.055} & \underline{0.080} & 0.061 & 0.088 & 0.058 & 0.083 & 0.071 & 0.096 & 0.070 & 0.102 \\ 
        ~ & 192 & \underline{0.058} & \textbf{0.085} & \textbf{0.057} & \textbf{0.085} & 0.061 & 0.088 & \underline{0.058} & \textbf{0.085} & 0.060 & 0.087 & 0.067 & 0.094 & 0.064 & 0.090 & 0.073 & 0.101 & 0.071 & 0.100 \\ 
        ~ & 336 & \underline{0.063} & \textbf{0.091} & \textbf{0.062} & \underline{0.092} & 0.065 & 0.093 & \underline{0.063} & \underline{0.092} & 0.065 & 0.094 & 0.074 & 0.101 & 0.069 & 0.095 & 0.085 & 0.110 & 0.072 & 0.104 \\ 
        ~ & 720 & \textbf{0.069} & \textbf{0.096} & \underline{0.070} & \underline{0.098} & 0.072 & 0.100 & \textbf{0.069} & \underline{0.098} & 0.071 & 0.100 & 0.082 & 0.111 & 0.073 & 0.104 & 0.096 & 0.122 & 0.075 & 0.110 \\  \hline
        \multirow{4}*{\rotatebox{90}{ETTm2}} & 96 & \textbf{0.029} & \textbf{0.041} & 0.033 & 0.051 & \underline{0.030} & \underline{0.042} & 0.032 & 0.044 & \textbf{0.029} & \textbf{0.041} & 0.065 & 0.083 & 0.038 & 0.056 & 0.031 & 0.043 & 0.039 & 0.058 \\ 
        ~ & 192 & \underline{0.033} & \underline{0.047} & 0.038 & 0.060 & \textbf{0.032} & \underline{0.047} & 0.036 & 0.049 & \textbf{0.032} & \textbf{0.046} & 0.079 & 0.101 & 0.043 & 0.064 & 0.037 & 0.048 & 0.043 & 0.065 \\ 
        ~ & 336 & \textbf{0.037} & \textbf{0.052} & 0.045 & 0.068 & \underline{0.038} & \underline{0.053} & 0.040 & 0.055 & 0.039 & 0.054 & 0.094 & 0.121 & 0.048 & 0.072 & 0.044 & 0.059 & 0.046 & 0.070 \\ 
        ~ & 720 & \textbf{0.041} & \underline{0.059} & 0.051 & 0.078 & \underline{0.043} & 0.061 & 0.044 & 0.061 & \textbf{0.041} & \textbf{0.058} & 0.128 & 0.160 & 0.055 & 0.082 & 0.060 & 0.084 & 0.054 & 0.081 \\ \hline
        \multirow{4}*{\rotatebox{90}{Weather}} & 96 & \textbf{0.038} & \textbf{0.076} & \textbf{0.038} & \underline{0.077} & \underline{0.039} & 0.080 & 0.040 & 0.083 & 0.041 & 0.081 & 0.045 & \underline{0.077} & 0.052 & 0.085 & 0.070 & 0.102 & 0.059 & 0.094 \\
        ~ & 192 & \textbf{0.045} & \textbf{0.085}  & \textbf{0.045} & \underline{0.087} & \underline{0.048} & 0.091 & 0.049 & 0.100 & \underline{0.048} & 0.090 & 0.052 & \textbf{0.085} & 0.059 & 0.095 & 0.086 & 0.124 & 0.066 & 0.105 \\
        ~ & 336 & \textbf{0.052} & \underline{0.096} & \underline{0.053} & 0.097 & 0.055 & 0.099 & 0.059 & 0.141 & 0.057 & 0.099 & 0.058 & \textbf{0.093} & 0.068 & 0.105 & 0.091 & 0.133 & 0.068 & 0.107 \\
        ~ & 720 & \textbf{0.060} & \underline{0.106} & \textbf{0.060} & 0.107 & 0.062 & 0.107 & \underline{0.063} & \underline{0.106} & 0.065 & 0.108 & 0.066 & \textbf{0.103} & 0.074 & 0.115 & 0.116 & 0.160 & 0.078 & 0.121 \\ \hline
        \multirow{4}*{\rotatebox{90}{Air}} & 96 & \textbf{0.058}  & \underline{0.089} & 0.060 & 0.095 & \underline{0.059} & 0.092 & 0.070 & 0.096 & 0.062 & \textbf{0.088} & 0.111 & 0.174 & 0.101 & 0.180 & 0.089 & 0.130 & 0.093 & 0.127 \\ 
        ~ & 192 & \textbf{0.061} & \textbf{0.093} & \textbf{0.061} & 0.098 & \textbf{0.061} & \underline{0.094} & 0.075 & 0.103 & \underline{0.070} & 0.096 & 0.120 & 0.187 & 0.108 & 0.190 & 0.108 & 0.150 & 0.116 & 0.191 \\ 
        ~ & 336 & \textbf{0.068}  & \underline{0.099} & \underline{0.069} & 0.104 & 0.070 & 0.105 & 0.075 & 0.103 & 0.071 & \textbf{0.097} & 0.126 & 0.194 & 0.112 & 0.193 & 0.143 & 0.189 & 0.120 & 0.201 \\ 
        ~ & 720 & \underline{0.077}  & \underline{0.110} & 0.083 & 0.127 & 0.082 & 0.119 & 0.084 & 0.113 & \textbf{0.076} & \textbf{0.102} & 0.138 & 0.209 & 0.129 & 0.211 & 0.186 & 0.272 & 0.125 & 0.205 \\ \hline
        \multirow{2}*{\rotatebox{90}{Num}} & First & 21 & 20 & 9 & 1 & 3 & 1 & 2 & 2 & 6 & 9 & 0 & 3 & 1 & 0 & 0 & 0 & 0 & 0 \\
        ~ & Second & 7 & 8 & 6 & 7 & 9 & 8 & 5 & 7 & 7 & 4 & 0 & 1 & 1 & 2 & 0 & 0 & 0 & 0 \\ \hline
    \end{tabular}}
    \label{tab: Main results}
    \vspace{-1.5em}
\end{table*}

\textbf{Baselines.} We select eight SOTA models as baselines, covering a diverse range of design paradigms for long-term time series forecasting (LTSF). These include time-domain models, frequency-domain models, and hybrid architectures. The full names and official implementations of these baselines are provided in Table \ref{tab: baselines}. We briefly categorize and describe them as follows:
\begin{itemize}
    \item \textbf{Time-domain methods.}  LTSF-Linear \cite{Zeng2022AreTE} is a lightweight, single-layer linear model designed to capture local temporal dependencies.  
TSMixer \cite{chen2023tsmixer} employs a fully MLP-based architecture that stacks multiple perceptrons to model complex temporal patterns. Informer \cite{zhou2021informer} is a Transformer-based model that utilizes ProbSparse attention and generative decoding to facilitate long-sequence forecasting.

    \item \textbf{Frequency-domain methods.}  
FreTS \cite{yi2024frequency} enhances global signal modeling by applying  MLPs to frequency-domain representations.  
FEDformer \cite{zhou2022fedformer} introduces a frequency-enhanced, decomposition-based Transformer architecture.

    \item \textbf{Hybrid/Dual-branch methods.}
Autoformer \cite{wu2021autoformer} is a Transformer variant that leverages auto-correlation and series decomposition to model temporal patterns.   
TimeKAN \cite{huang2025timekan} integrates KAN with frequency representations derived via FFT/IFFT but applies KAN exclusively in the time domain.
ATFNet \cite{ye2024atfnet} features a dual-branch design that combines time- and frequency-domain modules to capture both local and global dependencies.
\end{itemize}

\textbf{Metrics.} To evaluate the performance of TFKAN, two commonly used metrics, Mean Absolute Error (MAE) and Root Mean Squared Error (RMSE), are selected. MAE measures the average magnitude of errors, providing an intuitive interpretation of prediction accuracy. RMSE emphasizes larger errors, offering a more comprehensive evaluation of model robustness.

\textbf{Experimental Environment.}  
All code is implemented in Python using the PyTorch deep learning framework, developed within the PyCharm IDE on a Windows 10 environment. All experiments are conducted on a server running Ubuntu 20.04, equipped with an NVIDIA GeForce RTX 3090 GPU (24 GB VRAM), an Intel Xeon Gold 6230 CPU, and 128 GB of RAM. To ensure reproducibility, random seeds are fixed across NumPy, PyTorch, and Python’s built-in random module. The KAN module is configured with a grid size of $s=2$ and a spline order of $k=1$, with each KAN layer using a hidden size of 258. Input data is normalized using the MinMaxScaler. We adopt the Adam optimizer with a learning rate adjusted from the range $[10^{-6}, 10^{-2}]$, and batch sizes are adjusted from the range $[4, 64]$. All models are trained for up to 10 epochs with early stopping based on validation loss.

\subsection{Main Results} \label{sec: mainResults}

To address RQ1, we conduct a comprehensive evaluation of TFKAN against eight SOTA forecasting models on seven widely used time series datasets. The overall results are presented in Table \ref{tab: Main results}. Following standard practice \cite{wu2021autoformer}, we set the lookback window size to $L=96$ for all datasets. For the relatively short ILI dataset, prediction lengths are chosen as $\tau \in \{24,36,48,60\}$, while for the remaining datasets, we use $\tau \in \{96,192,336,720\}$.

\textbf{Overall Performance.}  
TFKAN achieves consistently superior results across all datasets and forecasting lengths, outperforming all baseline methods in terms of both MAE and RMSE. On the ILI dataset, TFKAN achieves the best performance across all prediction lengths. For example, at $\tau=24$, it achieves an MAE of 0.146 and an RMSE of 0.211. On ETTh2 with $\tau=720$, TFKAN achieves a MAE of 0.047 and RMSE of 0.065, significantly outperforming all other methods.

\textbf{Comparison with Frequency-domain and Hybrid Baselines.}  
To assess the benefits of frequency-domain modeling, we include several strong baselines that leverage spectral information, including Autoformer, FreTS, FEDformer, TimeKAN, and ATFNet. These models incorporate frequency components through various mechanisms such as spectral decomposition, energy-based MLP design, or FFT/IFFT processing. Among them, TimeKAN achieves competitive results on ETTh1, while ATFNet performs strongly on ETTh2 due to its dual-branch attention and extended DFT module. Nevertheless, TFKAN consistently surpasses all these frequency-domain and hybrid models, highlighting the advantage of directly applying KAN in the frequency domain. For instance, on ETTh2 with $\tau=336$, TFKAN achieves a MAE of 0.042 and RMSE of 0.059, outperforming ATFNet (0.067/0.087) and TimeKAN (0.058/0.083) by a significant margin.

\textbf{Considering Dataset Characteristics.}  
TFKAN demonstrates strong adaptability across datasets of different scales and characteristics. On smaller datasets like ILI, it shows excellent generalization and robustness. On large-scale datasets with longer sequences, such as Weather and ETTm1, TFKAN maintains SOTA performance by effectively capturing long-range dependencies and periodic structures.

\subsection{Ablation Studies} \label{sec: ablation}
To answer RQ2, a series of ablation studies is conducted to evaluate the effectiveness of the key components in TFKAN. Specifically, the contributions of the KAN-based modules, the dual-branch architecture, and the dimension adjustment strategy are analyzed.

\begin{table*}[!ht]
    \centering
    \caption{Ablation study comparing TFKAN with variants where two-layer MLPs partially or fully replace KAN modules. The lookback window size is fixed to $L = 96$. Metrics are MAE and RMSE. Best results are highlighted in bold.}
    \resizebox{0.99\textwidth}{!}{
    \begin{tabular}{cc|cc|cc|cc|cc|cc|cc|cc|cc}
    \hline
    \multicolumn{2}{c|}{Model} & \multicolumn{2}{c|}{TFKAN} & \multicolumn{2}{c|}{MLP} & \multicolumn{2}{c|}{MLP\_time} & \multicolumn{2}{c|}{MLP\_freq} & \multicolumn{2}{c|}{MLP\_pred} & \multicolumn{2}{c|}{MLP\_time\_freq} & \multicolumn{2}{c|}{MLP\_pred\_time} & \multicolumn{2}{c}{MLP\_pred\_freq} \\
    \multicolumn{2}{c|}{Metrics} & MAE & RMSE & MAE & RMSE & MAE & RMSE & MAE & RMSE & MAE & RMSE & MAE & RMSE & MAE & RMSE & MAE & RMSE \\ \hline
    \multirow{4}*{\rotatebox{90}{ILI}} & 24 & \textbf{0.146} & \textbf{0.211} & 0.152 & 0.217 & 0.152 & 0.216 & 0.152 & 0.217 & 0.159 & 0.220 & 0.157 & 0.222 & 0.162 & 0.225 & 0.163 & 0.227 \\
    ~&36 & \textbf{0.137} & \textbf{0.206} & 0.140 & 0.207 & 0.141 & 0.207 & 0.140 & 0.208 & 0.151 & 0.212 & 0.139 & 0.207 & 0.147 & 0.210 & 0.150 & 0.212 \\
    ~&48 & \textbf{0.139} & \textbf{0.206} & 0.141 & 0.207 & 0.143 & 0.208 & 0.141 & 0.208 & 0.157 & 0.220 & 0.148 & 0.213 & 0.153 & 0.217 & 0.165 & 0.228 \\
    ~&60 & \textbf{0.149} & \textbf{0.215} & 0.150 & 0.216 & 0.153 & 0.219 & 0.150 & 0.216 & 0.169 & 0.233 & 0.157 & 0.223 & 0.166 & 0.229 & 0.171 & 0.234 \\ \hline
    \multirow{4}*{\rotatebox{90}{ETTh1}} & 96 & \textbf{0.061} & \textbf{0.088} & 0.063 & 0.090 & 0.062 & 0.089 & 0.063 & 0.090 & 0.064 & 0.091 & 0.062 & 0.089 & 0.063 & 0.091 & 0.064 & 0.092 \\
    ~&192 & \textbf{0.067} & \textbf{0.093} & 0.068 & 0.094 & 0.069 & 0.094 & 0.068 & 0.094 & 0.070 & 0.097 & 0.068 & 0.094 & 0.070 & 0.097 & 0.070 & 0.097 \\
    ~&336 & \textbf{0.071} & \textbf{0.098} & 0.072 & 0.099 & 0.072 & 0.100 & 0.072 & 0.099 & 0.075 & 0.103 & 0.072 & 0.099 & 0.074 & 0.102 & 0.074 & 0.101 \\
    ~&720 & \textbf{0.082} & \textbf{0.109} & 0.084 & 0.111 & 0.083 & 0.110 & 0.084 & 0.111 & 0.087 & 0.115 & 0.083 & 0.110 & 0.087 & 0.115 & 0.085 & 0.113 \\ \hline
    \multirow{4}*{\rotatebox{90}{ETTh2}} & 96 & \textbf{0.036} & \textbf{0.052} & 0.038 & 0.053 & 0.038 & 0.054 & 0.038 & 0.053 & 0.038 & 0.053 & 0.038 & 0.053 & 0.039 & 0.055 & 0.037 & 0.053 \\
    ~&192 & \textbf{0.040} & \textbf{0.057} & 0.043 & 0.059 & 0.042 & 0.058 & 0.043 & 0.059 & 0.041 & 0.058 & 0.043 & 0.059 & 0.045 & 0.061 & 0.043 & 0.059 \\
    ~&336 & \textbf{0.041} & \textbf{0.058} & 0.047 & 0.064 & 0.047 & 0.064 & 0.047 & 0.064 & 0.046 & 0.062 & 0.050 & 0.066 & 0.047 & 0.064 & 0.047 & 0.065 \\
    ~&720 & \textbf{0.047} & \textbf{0.065} & 0.060 & 0.081 & 0.066 & 0.091 & 0.060 & 0.081 & 0.067 & 0.087 & 0.078 & 0.105 & 0.074 & 0.105 & 0.068 & 0.099 \\ \hline
    \multirow{4}*{\rotatebox{90}{Air}} & 96 & \textbf{0.058} & \textbf{0.089} & 0.063 & 0.091 & 0.060 & 0.090 & 0.063 & 0.091 & 0.062 & 0.092 & 0.061 & 0.091 & 0.063 & 0.092 & 0.064 & 0.094 \\
    ~&192 & \textbf{0.061} & \textbf{0.093} & 0.070 & 0.098 & 0.070 & 0.098 & 0.070 & 0.098 & 0.069 & 0.098 & 0.070 & 0.100 & 0.079 & 0.106 & 0.077 & 0.106 \\
    ~&336 & \textbf{0.068} & \textbf{0.099} & 0.079 & 0.108 & 0.072 & 0.101 & 0.079 & 0.108 & 0.073 & 0.102 & 0.077 & 0.107 & 0.094 & 0.120 & 0.095 & 0.121 \\
    ~&720 & \textbf{0.077} & \textbf{0.110} & 0.094 & 0.123 & 0.086 & 0.113 & 0.094 & 0.123 & 0.085 & 0.116 & 0.088 & 0.120 & 0.087 & 0.118 & 0.090 & 0.123 \\ \hline
    \end{tabular}}
    \label{tab: Ablation_mlp}
\end{table*}

\textbf{TFKAN vs. MLP.}
To investigate the contribution of KAN modules in different parts of the architecture, we conduct comprehensive ablation experiments by replacing KAN with standard two-layer MLPs. Specifically, we test the following configurations: MLP, where all KANs are replaced; MLP\_time, MLP\_freq, and MLP\_pred, where KANs are removed only from the time branch, frequency branch, or the predictor, respectively; and their combinations, such as MLP\_time\_freq, MLP\_pred\_freq, and MLP\_pred\_time. The results are reported in Table \ref{tab: Ablation_mlp}, evaluated on four datasets (ILI, ETTh1, ETTh2, and Air) across various forecasting lengths. All experiments use a fixed lookback window size of $L=96$ and are measured in terms of MAE and RMSE. \textbf{1) Consistent effectiveness of KAN.} TFKAN achieves the best performance across all datasets and prediction lengths in every case. For instance, on ILI with prediction length 24, TFKAN reaches a MAE of 0.146 and RMSE of 0.211, outperforming the fully MLP-based model (MAE: 0.152, RMSE: 0.217). The performance gap persists across larger prediction lengths and other datasets. \textbf{2) Each KAN branch independently contributes to performance.} Replacing KAN in any single module—whether in the MLP\_time, MLP\_freq, or the MLP\_pred, leads to consistent performance drops compared to the full TFKAN model. For instance, on the ETTh2 dataset with prediction length 720, MLP\_time, MLP\_freq, and MLP\_pred all show higher MAE and RMSE than TFKAN, increasing MAE from 0.047 to 0.066, 0.060, and 0.067, respectively. Similar trends are observed across other datasets. These results indicate that each KAN component plays a distinct role in capturing temporal or frequency dynamics, and none can be removed without compromising the model's accuracy. \textbf{3) Compound replacements often lead to larger performance degradation.} When KAN is removed from multiple components, such as in MLP\_pred\_time or MLP\_pred\_freq, performance tends to degrade more significantly compared to removing only a single component. For example, on the Air dataset with prediction length 336, MLP\_pred\_freq yields a MAE of 0.095, which is notably worse than MLP\_pred at 0.073. These prove the cumulative benefit of retaining KAN in multiple modules.

\begin{table}[!ht]
    \centering
    \caption{Comparison of TFKAN with single-branch architectures (Only\_Time and Only\_Freq) across four datasets. The evaluation is also based on MAE and RMSE, with a lookback window size of $L=96$. Best results are highlighted in bold.} 
    \resizebox{0.49\textwidth}{!}{
    \begin{tabular}{cc|cc|cc|cc}
    \hline
        \multicolumn{2}{c|}{Model} & \multicolumn{2}{c|}{TFKAN} & \multicolumn{2}{c|}{Only\_Time} & \multicolumn{2}{c}{Only\_Freq} \\
        \multicolumn{2}{c|}{Metrics} & MAE & RMSE & MAE & RMSE & MAE & RMSE \\ \hline
        \multirow{4}*{\rotatebox{90}{ILI}} & 24 & \textbf{0.146}  & \textbf{0.211}  & 0.226  & 0.296  & 0.169  & 0.236  \\ 
        ~ & 36 & \textbf{0.137}  & \textbf{0.206}  & 0.350  & 0.477  & 0.157  & 0.223  \\ 
        ~ & 48 & \textbf{0.139}  & \textbf{0.206}  & 0.375  & 0.514  & 0.153  & 0.219  \\ 
        ~ & 60 & \textbf{0.149}  & \textbf{0.215}  & 0.384  & 0.527  & 0.167  & 0.235  \\ \hline
        \multirow{4}*{\rotatebox{90}{ETTh1}} & 96 & \textbf{0.061}  & \textbf{0.088}  & 0.066  & 0.094  & 0.066  & 0.093  \\ 
        ~ & 192 & \textbf{0.067} & \textbf{0.093} & 0.071  & 0.098  & 0.069  & 0.096  \\ 
        ~ & 336 & \textbf{0.071} & \textbf{0.098} & 0.078  & 0.106  & 0.073  & 0.100  \\ 
        ~ & 720 & \textbf{0.082} & \textbf{0.109} & 0.093  & 0.121  & 0.087  & 0.115  \\ \hline
        \multirow{4}*{\rotatebox{90}{ETTh2}} & 96 & \textbf{0.036} & \textbf{0.052} & 0.040  & 0.055  & 0.041  & 0.057  \\ 
        ~ & 192 & \textbf{0.040} & \textbf{0.057} & 0.044  & 0.062  & 0.047  & 0.066  \\ 
        ~ & 336 & \textbf{0.041}  & \textbf{0.058}  & 0.049  & 0.066  & 0.054  & 0.080  \\ 
        ~ & 720 & \textbf{0.047} & \textbf{0.065} & 0.057  & 0.076  & 0.074  & 0.105  \\ \hline
        \multirow{4}*{\rotatebox{90}{Air}} & 96 & \textbf{0.058}  & \textbf{0.089}  & 0.068  & 0.101  & 0.061  & 0.090  \\ 
        ~ & 192 & \textbf{0.061}  & \textbf{0.093}  & 0.071  & 0.104  & 0.071  & 0.098  \\ 
        ~ & 336 & \textbf{0.068}  & \textbf{0.099}  & 0.075  & 0.108  & 0.074  & 0.104  \\ 
        ~ & 720 & \textbf{0.077}  & \textbf{0.110}  & 0.086  & 0.118  & 0.085  & 0.117 \\ \hline
    \end{tabular}}
    \label{tab: Ablation_dual}
\end{table}

\textbf{Dual-Branch vs. Single-Branch Architectures.}  
To evaluate the effectiveness of the dual-branch architecture in TFKAN, we compare it with two single-branch variants: Only\_Time and Only\_Freq. The Only\_Time variant retains only the time-domain branch, while Only\_Freq keeps only the frequency-domain branch. Table \ref{tab: Ablation_dual} summarizes the results across four datasets using the same experimental settings as in the KAN ablation. \textbf{1) Dual-branch TFKAN consistently outperforms single-branch variants.} TFKAN achieves the best performance in all datasets and prediction lengths. For example, on the ILI dataset with $\tau = 24$, TFKAN obtains a MAE of 0.146 and RMSE of 0.211, significantly outperforming Only\_Time (MAE: 0.226, RMSE: 0.296) and Only\_Freq (MAE: 0.169, RMSE: 0.236). This consistent trend holds across longer prediction lengths and other datasets. \textbf{2) Time-frequency integration improves generalization.} Both single-branch models exhibit inferior performance compared to the full TFKAN, indicating that time-domain and frequency-domain information are complementary. Notably, on ETTh2 with $\tau = 720$, TFKAN achieves a MAE of 0.047 and RMSE of 0.065, while Only\_Time and Only\_Freq degrade to 0.057 / 0.076 and 0.074 / 0.105, respectively. These results support that integrating both domains enables more robust and generalizable forecasting, and thus validates the architectural design of TFKAN.

\begin{table}[!h]
    \centering
    \caption{Comparison of TFKAN with three dimension adjustment variants (All\_Adjust, No\_Adjust, and Only\_Time\_Adjust) across four datasets. The evaluation is based on MAE and RMSE with a lookback window size of $L=96$. Best results are highlighted in bold, while the second-best results are underlined.}
    \resizebox{0.49\textwidth}{!}{
    \begin{tabular}{cc|cc|cc|cc|cc}
    \hline
        \multicolumn{2}{c|}{Model} & \multicolumn{2}{c|}{TFKAN} & \multicolumn{2}{c|}{All\_Adjust} & \multicolumn{2}{c|}{No\_Adjust} & \multicolumn{2}{c}{Only\_Time\_Adjust} \\
        \multicolumn{2}{c|}{Metrics} & MAE & RMSE & MAE & RMSE & MAE & RMSE & MAE & RMSE \\ \hline
        \multirow{4}*{\rotatebox{90}{ILI}} & 24 & \textbf{0.146}  & \textbf{0.211}  & \underline{0.159}  & \underline{0.223}  & 0.219  & 0.292  & 0.226  & 0.296  \\ 
        ~ & 36 & \textbf{0.137}  & \textbf{0.206}  & \underline{0.148}  & \underline{0.214}  & 0.267  & 0.366  & 0.350  & 0.477  \\ 
        ~ & 48 & \textbf{0.139}  & \textbf{0.206}  & \underline{0.153} & \underline{0.219}  & 0.287  & 0.395  & 0.375  & 0.514  \\ 
        ~ & 60 & \textbf{0.149}  & \textbf{0.215}  & \underline{0.164}  & \underline{0.230}  & 0.296  & 0.406  & 0.384  & 0.527  \\ \hline
        \multirow{4}*{\rotatebox{90}{ETTh1}} & 96 & \textbf{0.061}  & \textbf{0.088}  & \underline{0.063}  & \underline{0.091}  & 0.066  & 0.093  & 0.066  & 0.094  \\ 
        ~ & 192 & \textbf{0.067} & \textbf{0.093} & \underline{0.069}  & \underline{0.096}  & 0.074  & 0.101  & 0.071  & 0.098  \\ 
        ~ & 336 & \textbf{0.071} & \textbf{0.098} & \underline{0.072}  & \underline{0.099}  & 0.074  & 0.102  & 0.078  & 0.106  \\ 
        ~ & 720 & \textbf{0.082} & \textbf{0.109} & \underline{0.085}  & \underline{0.112}  & 0.093  & 0.120  & 0.093  & 0.121  \\ \hline
        \multirow{4}*{\rotatebox{90}{ETTh2}} & 96 & \textbf{0.036} & \textbf{0.052} & 0.040  & 0.056  & 0.041  & 0.057  & \underline{0.038}  & \underline{0.054}  \\ 
        ~ & 192 & \textbf{0.040} & \textbf{0.057} & \underline{0.042}  & \underline{0.059}  & 0.049  & 0.068  & 0.044  & 0.062  \\ 
        ~ & 336 & \textbf{0.041}  & \textbf{0.058}  & \underline{0.047}  & \underline{0.066}  & 0.057  & 0.080  & 0.049  & \underline{0.066}  \\ 
        ~ & 720 & \textbf{0.047} & \textbf{0.065} & \underline{0.056}  & \underline{0.075}  & 0.058  & 0.077  & 0.057  & 0.076  \\ \hline
        \multirow{4}*{\rotatebox{90}{Air}} & 96 & \textbf{0.058}  & \textbf{0.089}  & 0.064  & \underline{0.091}  & \underline{0.063}  & 0.093  & 0.068  & 0.101  \\ 
        ~ & 192 & \textbf{0.061}  & \textbf{0.093}  & 0.068  & \underline{0.097}  & \underline{0.066}  & \underline{0.097}  & 0.071  & 0.104  \\ 
        ~ & 336 & \textbf{0.068}  & \textbf{0.099}  & \underline{0.074}  & \underline{0.103}  & 0.078  & 0.109  & 0.075  & 0.108  \\ 
        ~ & 720 & \textbf{0.077}  & \textbf{0.110}  & 0.091  & 0.126  & 0.086  & 0.118  & \underline{0.083}  & \underline{0.116} \\ \hline
    \end{tabular}}
    \label{tab: Ablation_dimension}
\end{table}

\textbf{Dimension Adjustment Strategy.}  
To evaluate the effectiveness of the proposed dimension adjustment mechanism in TFKAN, we compare the full model against three ablated variants: All\_Adjust, No\_Adjust, and Only\_Time\_Adjust. Specifically, All\_Adjust applies upscaling to all hidden dimensions in both time and frequency branches, No\_Adjust omits dimension adjustment entirely, and Only\_Time\_Adjust applies the adjustment only in the time-domain branch. Table \ref{tab: Ablation_dimension} summarizes the results across four datasets under the same settings as previous ablations. \textbf{1) TFKAN consistently outperforms all variants.} Across all datasets and prediction lengths, TFKAN achieves the lowest MAE and RMSE values, as highlighted in bold. For example, on the ILI dataset with $\tau=24$, TFKAN obtains a MAE of 0.146 and RMSE of 0.211, outperforming All\_Adjust (0.159 / 0.223), No\_Adjust (0.219 / 0.292), and Only\_Time\_Adjust (0.226 / 0.296). These results confirm the effectiveness of our selective dimension adjustment strategy. \textbf{2) Selective adjustment is more effective than full or no adjustment.}  Among the ablated models, All\_Adjust generally performs better than both No\_Adjust and Only\_Time\_Adjust, but still underperforms TFKAN. This suggests that indiscriminately increasing dimensionality across all branches may introduce redundancy or noise, limiting performance.

\begin{table}[!ht]
    \centering
    \caption{Comparison of TFKAN with two separate KAN (Two\_FreqKAN) for processing the real and imaginary components in the frequency branch. Results are reported on four datasets with multiple prediction lengths (lookback window $L=96$). The best performance for each setting is highlighted in bold.}
    \begin{tabular}{cc|cc|cc}
    \hline
        \multicolumn{2}{c|}{Model} & \multicolumn{2}{c|}{TFKAN} & \multicolumn{2}{c}{Two\_FreqKAN} \\
        \multicolumn{2}{c|}{Metrics} & MAE & RMSE & MAE & RMSE \\  \hline
        \multirow{4}*{\rotatebox{90}{ILI}} & 24 & 0.146 & 0.211 & \textbf{0.140} & \textbf{0.207} \\ 
        ~ & 36 & \textbf{0.137} & 0.206 & \textbf{0.137} & \textbf{0.205} \\ 
        ~ & 48 & \textbf{0.139} & 0.206 & \textbf{0.139} & \textbf{0.204} \\ 
        ~ & 60 & \textbf{0.149} & \textbf{0.215} & 0.154 & 0.220 \\ \hline
        \multirow{4}*{\rotatebox{90}{ETTh1}} & 96 & \textbf{0.061} & \textbf{0.088} & \textbf{0.061} & \textbf{0.088} \\ 
        ~ & 192 & 0.067 & \textbf{0.093} & \textbf{0.066} & \textbf{0.093} \\ 
        ~ & 336 & \textbf{0.071} & \textbf{0.098} & \textbf{0.071} & \textbf{0.098} \\ 
        ~ & 720 & \textbf{0.082} & \textbf{0.109} & \textbf{0.082} & \textbf{0.109} \\ \hline
        \multirow{4}*{\rotatebox{90}{ETTh2}} & 96 & \textbf{0.036} & \textbf{0.052} & 0.038 & 0.053 \\ 
        ~ & 192 & \textbf{0.040} & \textbf{0.057} & \textbf{0.040} & \textbf{0.057} \\ 
        ~ & 336 & \textbf{0.041} & \textbf{0.058} & 0.042 & 0.060 \\ 
        ~ & 720 & \textbf{0.047} & \textbf{0.065} & 0.058 & 0.077 \\ \hline
        \multirow{4}*{\rotatebox{90}{Air}} & 96 & \textbf{0.058} & \textbf{0.089} & 0.059 & 0.090 \\ 
        ~ & 192 & \textbf{0.061} & \textbf{0.093} & 0.066 & 0.101 \\ 
        ~ & 336 & \textbf{0.068} & \textbf{0.099} & 0.081 & 0.111 \\ 
        ~ & 720 & \textbf{0.077} & \textbf{0.110} & 0.078 & 0.111 \\ \hline
    \end{tabular}
    \label{tab: Ablation_freq_kan}
\end{table}

\textbf{Shared vs. Independent Frequency-Domain KANs.}  
To explore whether using separate KAN modules for the real and imaginary parts of the frequency branch improves performance, we compare the standard TFKAN design, which uses a single shared KAN to process both parts, with a variant named Two\_FreqKAN, where two independent KANs are used. Table \ref{tab: Ablation_freq_kan} reports the results across four datasets. \textbf{1) Comparable performance on most datasets.} The overall results show that Two\_FreqKAN yields similar or slightly improved performance on some datasets such as ILI and ETTh1, e.g., on ILI ($\tau=24$), the MAE decreases from 0.146 to 0.140. However, in most cases, the differences are marginal. \textbf{2) Minor gains do not justify complexity.} While using two KANs for the frequency branch can improve expressiveness, it doubles the parameter count and computational cost for that component. On ETTh1, both variants perform almost identically. For example, on ETTh1 ($\tau=96$), both achieve MAE/RMSE of 0.061/0.088. Additionally, in the Fourier spectrum, the real and imaginary parts' local peaks are in the same position on the frequency axis, but differ only in amplitude/phase. The shared spline basis forces them to adjust synchronically at the same node, which is equivalent to imposing a "conjugate consistency" regularity on the network, which can suppress overfitting. Given the negligible accuracy gains and higher complexity, we retain the shared KAN design in TFKAN for its simplicity and efficiency.

\subsection{Parameter Sensitivity Analysis} \label{sec: sensitivity}
To address RQ3, parameter sensitivity studies are conducted to analyze the impact of two key hyperparameters: the input length $L$ and the embedding size $d$. These experiments evaluate how changes in $L$ and $d$ influence forecasting accuracy on prediction length $\tau=96$ (for ILI dataset, the $\tau=24$), measured by MAE and RMSE, and training efficiency, measured by average time per epoch.

\begin{figure}[htbp]
    \centering
    \includegraphics[width=1\linewidth]{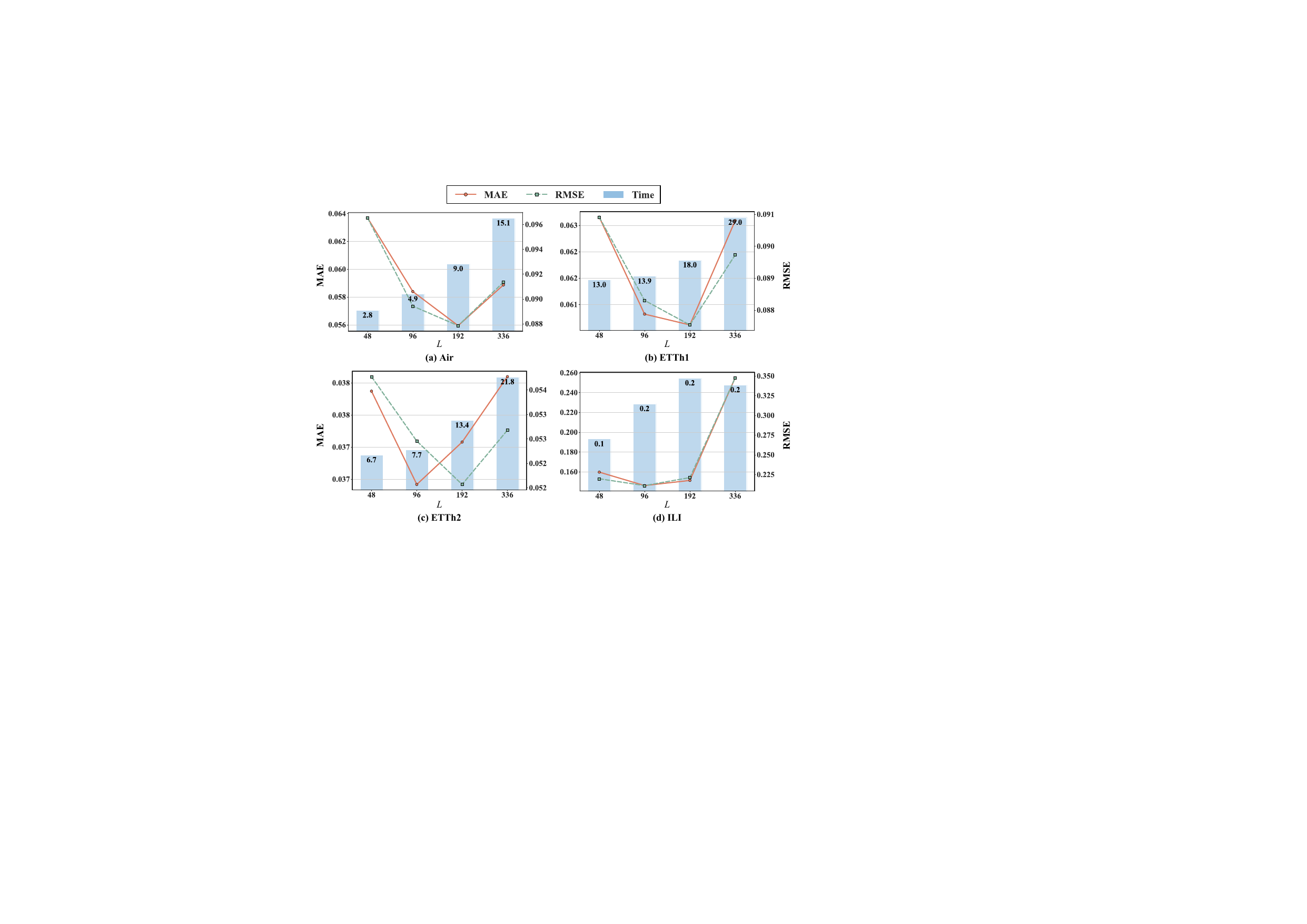}
    \caption{Parameter sensitivity studies of the input length $L$.}
    \label{fig: Sensitive_L}
\end{figure}

\textbf{Input Length $L$.}
The parameter sensitivity study on input length $L$ evaluates its effect on performance across four datasets: ETTh1, ETTh2, ILI, and Air. The results are visualized in Fig. \ref{fig: Sensitive_L}. The x-axis represents the input length $L$, while the orange line, green line, and blue bars represent MSE, RMSE, and average time per epoch, respectively. \textbf{1) Performance trends:}  Increasing $L$ generally enhances forecasting accuracy for most datasets, as it provides the model with more historical context. However, excessive input length can disrupt performance, possibly due to information overload. For instance, on the Air dataset, MAE improves from 0.064 to 0.056 when $L$ increases from 48 to 192, but a slight decline is observed at $L=336$. \textbf{2) Dataset characteristics:} On smaller datasets like ILI, increasing $L$ beyond 96 offers minimal accuracy improvements, while the additional computational overhead is negligible. However, MAE increases from 0.146 at $L=96$ to 0.255 at $L=336$, indicating diminishing returns for larger input lengths. \textbf{3) Efficiency:} Larger input lengths increase computation time significantly. For example, on ETTh1, the time per epoch rises from 12.96 seconds at $L=48$ to 28.96 seconds at $L=336$. A balance between performance and efficiency is achieved at moderate $L$ values, such as $L = 96$ or $L=192$.

\begin{figure}[htbp]
    \centering
    \includegraphics[width=1\linewidth]{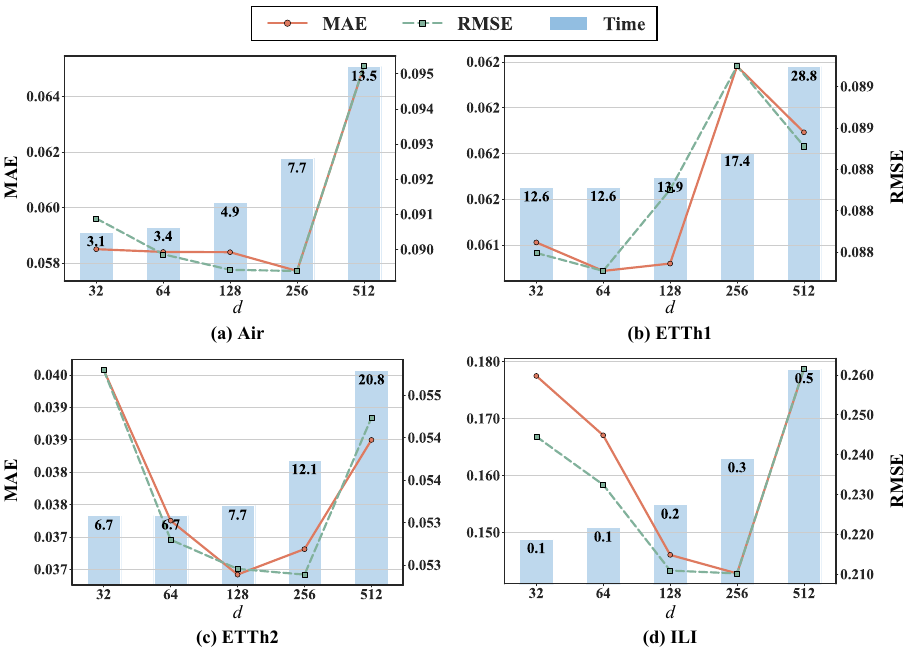}
    \caption{Parameter sensitivity studies of the embedding size $d$.}
    \label{fig: Sensitive_d}
\end{figure}

\textbf{Embedding Size $d$.}
The sensitivity study on embedding size $d$ evaluates its effect on model performance and training efficiency across the same datasets. The results are visualized in Fig. \ref{fig: Sensitive_d}. The x-axis represents embedding size $d$, while the orange, green, and blue lines indicate MSE, RMSE, and time per epoch, respectively. \textbf{1) Performance trends:} Increasing $d$ generally improves accuracy until a certain point, after which performance plateaus or slightly degrades. For example, on the ILI dataset, MAE decreases from 0.177 at $d=32$ to 0.146 at $d=128$, but then increases to 0.179 at $d=512$. Similarly, for the Air dataset, MAE increases to 0.065 at $d=512$. \textbf{2) Dataset characteristics:} Smaller datasets, such as ILI, are more sensitive to variations in $d$, with performance gains degrading as $d$ surpasses the optimal value. In contrast, larger datasets, such as ETTh1 and ETTh2, exhibit more stable performance across a range of $d$ values, with the changes within 0.002, demonstrating greater robustness to changes in this hyperparameter $d$. \textbf{3) Efficiency:} Larger embedding sizes significantly increase training times without corresponding improvements in accuracy. For instance, on ETTh2, the time per epoch increases from 7.67 seconds at $d=128$ to 20.77 seconds at $d=512$. As a result, an embedding size of $d=128$ strikes a balance between accuracy and efficiency for most datasets.

\subsection{Efficiency Analysis}  \label{sec: efficiency}
Table \ref{tab: efficiency} presents the parameter counts and GPU memory usage (under batch size = 64) for TFKAN and eight baseline models on the ILI dataset under different prediction lengths. Despite its strong performance, TFKAN maintains a relatively low memory footprint (approximately 251 MB across all settings), which is significantly lower than Transformer-based models like FEDformer, Informer, and Autoformer, whose GPU usage exceeds 1.5 GB in most cases. In terms of parameter count, TFKAN (16.3M) is comparable to FEDformer (14.5M–15.6M) and notably smaller than ATFNet (up to 19.4M). Compared to lightweight models such as LTSF-Linear and TSMixer, which have only a few thousand to a few hundred thousand parameters, TFKAN trades off slightly higher complexity for substantially better accuracy, as shown in previous sections. Overall, these results highlight that TFKAN strikes a favorable balance between forecasting performance and computational efficiency. Its GPU memory usage remains moderate while delivering SOTA accuracy across datasets and prediction lengths.

\begin{table*}[!ht]
\centering
\caption{Comparison of parameter count and GPU memory usage (batch size = 64) for all models on the ILI dataset under different prediction lengths.}
\begin{tabular}{c|c|cccccccc}
\hline
\textbf{Pred Len} & \textbf{TFKAN (Our)} & \textbf{TimeKAN} & \textbf{ATFNet} & \textbf{FreTS} & \textbf{LTSF-Linear} \\ \hline
~ & \multicolumn{5}{c}{\textit{Param Count / GPU Memory (MB, batch size = 64)}} \\
\hline
24 & 16.33M / 250.65 & 31.1K / 550.99 & 19.20M / 1718.67 & 3.22M / 283.39 & 4.66K / 1.53 \\
36 & 16.35M / 250.96 & 32.3K / 551.05 & 19.28M / 1720.12 & 3.22M / 283.53 & 6.98K / 1.96  \\
48 & 16.36M / 251.27 & 33.5K / 551.11 & 19.37M / 1719.97 & 3.22M / 283.67 & 9.31K / 2.39  \\
60 & 16.38M / 251.58 & 34.6K / 551.17 & 19.45M / 1721.78 & 3.22M / 283.81 & 11.6K / 2.82 \\
\hline
\textbf{Pred Len} & \textbf{TFKAN (Our)} & \textbf{TSMixer} & \textbf{FEDformer} & \textbf{Informer} & \textbf{Autoformer} \\ \hline
~ & \multicolumn{5}{c}{\textit{Param Count / GPU Memory (MB, batch size = 64)}} \\
\hline
24 & 16.33M / 250.65 & 215.5K / 54.66 & 14.47M / 1508.79 & 11.33M / 1475.51 & 10.54M / 2019.37 \\
36 & 16.35M / 250.96 & 216.7K / 54.77 & 14.86M / 1575.80 & 11.33M / 1535.33 & 10.54M / 2139.58 \\
48 & 16.36M / 251.27 & 217.9K / 54.88 & 15.25M / 1641.18 & 11.33M / 1596.93 & 10.54M / 2290.05 \\
60 & 16.38M / 251.58 & 219.0K / 55.00 & 15.65M / 1713.07 & 11.33M / 1664.20 & 10.54M / 2460.84 \\
\hline
\end{tabular}
\label{tab: efficiency}
\end{table*}

\subsection{Predictions Visualizations}
\begin{figure}[htbp]
	\centering
        \includegraphics[width=1\linewidth]{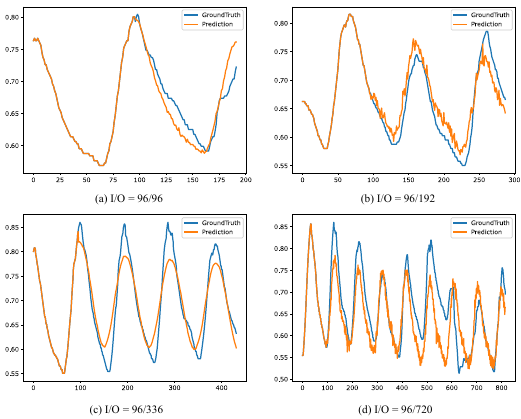}
	\caption{Visualizations of predictions (ground truth vs. predictions) on the ETTm2 dataset. ’I/O’ denotes lookback window size $L$ / prediction length $\tau$.}
	\label{fig: predictions}
\end{figure}

Fig. \ref{fig: predictions} visualizes the predictions and their comparison with the ground truth on the ETTm2 dataset. The 'I/O' notation refers to the input and output, representing the lookback window size $L$ and prediction length $\tau$, respectively.

\section{Conclusion}\label{sec:conclusion}

In this paper, we introduce a novel frequency-domain Kolmogorov-Arnold Network (KAN) for time series forecasting. Building upon this foundation, we propose the Time-Frequency Kolmogorov-Arnold Network (TFKAN). TFKAN employs a dual-branch architecture, where the frequency-domain KAN extracts frequency features and the time-domain KAN captures temporal dependencies, enabling the effective integration of complementary information from both domains. To address the heterogeneity between the time and frequency branches, we introduce a dimension-adjustment mechanism, which enhances the model's performance and robustness. Extensive experiments on seven time-series datasets demonstrate that TFKAN consistently outperforms eight SOTA methods, underscoring its superior forecasting capabilities. This work represents the first successful application of the frequency-domain KAN to time series forecasting. Future research will focus on optimizing TFKAN’s efficiency, particularly by reducing its parameter count, to further enhance its performance.

\bibliographystyle{IEEEtran}
\bibliography{refs}


 




\vfill

\end{document}